\documentclass[sigconf]{acmart}

\def\BibTeX{{\rm B\kern-.05em{\sc i\kern-.025em b}\kern-.08emT\kern-.1667em\lower.7ex\hbox{E}\kern-.125emX}}

\definecolor[named]{kellygreen}{cmyk}{.59, 0 .88, .27}
\usepackage[normalem]{ulem}

\copyrightyear{2020}
\acmYear{2020}
\setcopyright{acmlicensed}
\acmConference[FDG '20]{Foundations of Digital Games '20}{September 15--18, 2020}{Bugibba, Malta}
\acmBooktitle{Proceedings of FDG '20, September 15--18, 2020, Bugibba, Malta}
\acmPrice{}
\acmDOI{}
\acmISBN{}

\begin{document}
	
% The "title" command has an optional parameter, allowing the author to define a "short title" to be used in page headers.
%\title{How to Start Modeling Players before Data Gathering}
\title{Player Modeling via Multi-Armed Bandits}

%Player Modeling based on Social Comparison Orientation using Multi-armed Bandits 
%Title Keywords: Multi-Armed Bandit, Experience Management, Player Modeling, Social Comparison
% Multiple Regression Techniques for Player Modeling using Short-Horizon Multi-Armed Bandits

% bootstrap

% The "author" command and its associated commands are used to define the authors and their affiliations.
% Of note is the shared affiliation of the first two authors, and the "authornote" and "authornotemark" commands
% used to denote shared contribution to the research.
\author{Robert C. Gray}
\affiliation{\institution{Drexel University}}
\email{robert.c.gray@drexel.edu}
\author{Jichen Zhu}
\affiliation{\institution{Drexel University}}
\email{jichen.zhu@gmail.com}
\author{Danielle Arigo}
\affiliation{\institution{Rowan University}}
\email{arigo@rowan.edu}
\author{Evan Forman}
\affiliation{\institution{Drexel University}}
\email{emf27@drexel.edu}
\author{Santiago Onta\~n\'on}
\affiliation{\institution{Drexel University}}
\email{so367@drexel.edu}

% By default, the full list of authors will be used in the page headers. Often, this list is too long, and will overlap
% other information printed in the page headers. This command allows the author to define a more concise list
% of authors' names for this purpose.
% \renewcommand{\shortauthors}{Gray et al.}

% The abstract is a short summary of the work to be presented in the article.
\begin{abstract}
	This paper focuses on building personalized player models solely from player behavior in the context of adaptive games. We present two main contributions: The first is a novel approach to player modeling based on {\em multi-armed bandits} (MABs). This approach addresses, at the same time and in a principled way, both the problem of collecting data to model the characteristics of interest for the current player and the problem of adapting the interactive experience based on this model. Second, we present an approach to evaluating and fine-tuning these algorithms prior to generating data in a user study. This is an important problem, because conducting user studies is an expensive and labor-intensive process; therefore, an ability to evaluate the algorithms beforehand can save a significant amount of resources. We evaluate our approach in the context of modeling players' {\em social comparison orientation} (SCO) and present empirical results from both simulations and real players.
\end{abstract}


% Keywords. The author(s) should pick words that accurately describe the work being presented. Separate the keywords with commas.

% This command processes the author and affiliation and title information and builds the first part of the formatted document.
\maketitle

\section{Introduction}

Player modeling focuses on modeling and predicting player characteristics of interest, such as preferences, skill level, or behavior~\cite{Yannakakis2013}. One of the reasons player modeling is interesting is because it plays a key role in the creation of adaptive games. In this paper, we present two main contributions to player modeling: (1) a novel player modeling approach based on {\em multi-armed bandits} (MABs), and (2) an approach for evaluating and fine-tuning these algorithms before having access to real player data in a user study.

A common approach to player modeling is the use of machine learning (ML)~\cite{Drachen2009,valls2015exploring}; however, ML algorithms typically require large amounts of training data. 
%This is specially important if we are interested in training a personalized model for the player at hand, as we cannot expect the player to keep playing for an arbitrarily long amount of time until we have gathered enough data as for modeling her accurately. Another problem is how to use the player model to adapt the game experience, which is often handled separately (as done in standard {\em experience management} approaches~\cite{weyhrauch1997guiding, nelson2005search, sharma2010drama, thue2018toward}).
Our proposed approach to player modeling based on MABs~\cite{auer2002finite} solves both (1) this problem of training data acquisition as well as (2) the problem of how to use the player model to adapt the game.

%A multi-armed bandit problem is a sequential decision problem where an agent needs to iteratively choose one among a series of predefined options, obtaining a reward after each choice. The goal of the agent is to maximize the cumulative reward. 
We address the challenge of adapting a game to achieve a desired effect on the player by periodically choosing one among many potential ways to adapt the game. After observing the player's behavior in response to the bandit's choice, a reward value is generated based on the efficacy of that choice, which the bandit observes. We assume a lack of any prior training data before the user starts interacting with the system (though pre-existing data can be exploited). MAB strategies naturally solve this problem by balancing {\em exploration} (i.e., trying new ways to adapt the game to improve its understanding of the player) and {\em exploitation} (i.e., adapting the game in ways that have proved to work well in the past). %MABs are thus a natural fit for player modeling in adaptive games which, to the best of our knowledge have not been explored in the game AI literature.

Moreover, a second problem needs to be solved to effectively deploy this strategy, which constitutes our second contribution. Consider the common problem of choosing the appropriate AI approach before performing an adaptive game user study. How can we gain insights into which AI approaches would be best suited for our user study before engaging in the resource-intensive activity of actually carrying out the study? Additionally, how do we design the parameters of the user study (e.g., participants, duration) without knowing how the AI will perform? 
%
%Moreover, given the numerous applications of MABs, a large number of MAB strategies have been proposed in the literature, many of which have free parameters that need to be tuned to suit the specific application at hand. This raises a problem that is commonly faced by games researchers and designers when creating adaptive games: how can we ensure that the AI module's design is appropriately tuned for the specific task prior to conducting user studies that would evaluate the system's design? User studies are time and resource intensive, and thus ideally we would like to avoid having to run preliminary studies just with the purpose of comparing several AI techniques. Moreover, in an interventional study, the choices made by the AI over the course of the experiment will have a bearing on the observations generated. Thus, ensuring that the AI design is adequate for the study is an essential, but difficult task.
%
To solve this problem, we leverage publicly available data to create simulated players that exhibit statistical behavior patterns close to actual humans. Through the use case of modeling social comparison orientation (SCO) to maximize motivation toward physical activity, we show the promise of our approach as well as the effectiveness of our simulated players to evaluate MAB algorithms.

% - outline: finally, the last paragraph of an introduction is always the same, something like â€œthe remainder of this paper is structured as follows. First we present 
In the remainder of this paper, we first present some background on player modeling, adaptive games, and MABs. We then present our MAB player modeling framework, followed by our methodology for creating simulated players. Finally, we present empirical results from simulations (with the simulated players) and a real user study.

%The remainder of this paper is structured as follows: First, we present some background on player modeling, adaptive games, and multi-armed bandits (MAB). Then, we present our approach toward implementing an MAB-based player model of players' social comparison orientation (SCO) to motivate them to exercise. Third, we discuss our approach to create and use simulated players to evaluate MAB strategies. Finally, we examine the results of both our simulator experiments and the subsequent user study that deployed our solution.

\section{Background and Related Work}

This section briefly introduces some basic concepts of player modeling, adaptive games, multi-armed bandits, and simulated player-based evaluations.

\subsection{Player Modeling and Adaptive Games}

% Drop EM and replace with some high-level discussions on personalized adaptive games

% Explain adaptive game systems and how they rely on player models
Adaptive games leverage knowledge of the player to automatically adapt to better serve specific users or specific design goals~\cite{weyhrauch1997guiding,bates1992virtual,riedl2008dynamic,thue2018toward,zhu2019experience}. These methods often rely on player modeling to detect or predict a set of characteristics of the player that can inform the AI's decisions~\cite{Yannakakis2013}. Previous work has shown applications in improving learning outcomes~\cite{Valls2015} and health outcomes~\cite{zhu2018towards}, adjusting game difficulty~\cite{Alexander2013, Gray2018}, managing user interfaces~\cite{Gajos2017}, or even adapting game narratives~\cite{Sharma2010, Mateas2003, weyhrauch1997guiding}.

% Discuss how these approaches require an accurate starting point or quick convergence of the model with real users
%For an adaptive game to work effectively, it relies on the accuracy of its player model. Properly defining this model and maintaining its accuracy throughout the experience remains a central challenge in adaptive games research, and this challenge differs depending on the methods used to construct the model. 
Player models are often designed to leverage either {\it a priori} theory and heuristics (``top-down'') or assumption-free statistical methods (``bottom-up'') to perform their task of differentiating or defining players, where this dichotomy has also been suggested to define a spectrum \cite{Yannakakis2013}.
Our work is informed by both of these approaches; though our application domain is derived from psychology theory, we also wish to leverage the statistical power of context-agnostic modeling in the form of an MAB strategy. %Therefore, we propose a combination of these two approaches. Though we ultimately construct a bottom-up model for classifying players using machine learning techniques, this statistics-based model is trained on simulated players that implement a top-down model founded in theory.
%
% Explain our contribution - proposing a model for SCO via MAB classification (non-heuristic)
Specifically, we based our work on the psychology theory of social comparison (see Section \ref{section:modelingSCO}). Though heuristic-based models exist for classifying users based on their social comparison tendencies \cite{Gibbons1999}, our work employs a bottom-up approach using MAB strategies.

\subsection{Multi-Armed Bandits}

% Explain bandits at a general level
A multi-armed bandit (MAB) problem~\cite{auer2002finite} is a class of sequential decision problem where an agent needs to iteratively choose one among $k$ actions (called {\em arms}), after which it receives a stochastic reward. This mirrors the problem faced by a player in a casino deciding on which of the different gambling machines each of their tokens should be spent. %The agent needs to estimate the potential rewards of each action, balancing exploration and exploitation. %A MAB strategy is an algorithm to choose actions (arms) at each iteration.
The goal of an MAB strategy is to balance exploration and exploitation, assimilating new knowledge from rewards to ``converge'' on the arm with the maximum expected return as quickly as possible. 
Popular MAB strategies include the $\epsilon$-greedy strategy, where the arm that has historically returned the highest reward is always selected except in a portion of iterations (designated by $\epsilon$) where a random arm is chosen. Another is UCB1~\cite{auer2002finite}, which considers the upper confidence bound of expected rewards.

MAB strategies are interesting for adaptive games if we consider the different adaptation options as the arms in an MAB. 
%Such a strategy could then exploit the large amount of work done in this field to automatically adapt to specific users. 
Investigations into this have already begun in the context of adaptive interventions such as those that promote behavior change~\cite{forman2019can} with promising results. 
% Explain MAB strategies and how they differ to provide reasoning for why our simulation is needed
% Also introduce CMAB and contextual variations, which we explore as part of our contribution
However, to the best of our knowledge, MABs have not been used in the context of adaptive games or player modeling.

Applying MABs to player modeling raises an important challenge, however. There is a very large collection of MAB strategies proposed in the literature, each with their own practical and theoretical properties. These strategies are usually evaluated by their behavior ``in the limit'' (i.e., with large numbers of interactions with the environment). However, in a player modeling situation, we cannot expect the system to interact with players for this long. %For example, we cannot expect user studies where players interact with a system daily to exceed a few weeks, or at most a few months duration. 
This means that MAB strategies that work with very few interactions with users are needed, and thus we had to design an MAB strategy that satisfies these constraints before carrying out the study. %In the particular study that is the focus of this paper, we used MABs to choose one among several possible types of interventions day after day in our study, trying to choose the interventions that maximize physical activity (motivation). 
%A key challenge in designing the user study presented in this paper was how to design the MAB strategy prior to the study that would be effective with the expected duration of the study.

Therefore, our work pushes the state of the art in two separate ways. First, we present a novel player modeling framework based on MAB strategies. Second, we present an approach for evaluating strategies via simulated players to design an MAB strategy that is effective with very few player interactions.

%It is thus challenging to determine which strategy is the best for the problem at hand beforehand. %Additional variations explored in our research included Combinatorial MABs (CMABs)~\cite{???} that construct aggregate arms from sub-strategies, or Contextual MABs that consider outside information (context) other than arm rewards when making their choices.

% Explain how we are using MAB strategies

% In the case of this research, the MAB's task of converging on the optimal arm might be recast as such a ``bottom up'' method for constructing a model. In balancing exploration and exploitation to find the arm with the highest expected value in the most efficient way possible, it is ultimately seeking to determine among a set of defined options the one that best meets a researcher-defined criteria. In our research, we propose that if a problem of player modeling classification were established with the same structure of arms representing potential player types and rewards that represented {\it model affinity} (i.e., appropriateness of fit), the process of MAB convergence could be leveraged to instead determine the proper placement of a player within the model.

\subsection{AI Tuning via Simulation}

% Bias-variance tradeoff and the role (and considerations) for simulated users in evaluations
Carterette et al. \cite{Carterette2011} present a conceptualization of system evaluations as a continuum between {\em systems-based} approaches involving automated tests that evaluate predetermined scenarios and {\em user studies} involving real user interactions with the system. The former are viewed to have the advantages of stability, repeatability, and low costs at the risk of oversimplifying assumptions that could invalidate results. The latter are capable of answering more questions with potentially higher accuracy, but in exchange they carry a burden of higher expense and variability. This is referred to as the {\em bias-variance tradeoff} \cite{Carterette2011}.

For adaptive games and player modeling, it is difficult to escape the requirement of genuine user studies; however, researchers have found value in simulations for a number of situations. These may include the rarity of real players \cite{Chen2017}, the complexity of the test space \cite{Wendel2014}, a desire to maintain specific control over how a model is trained \cite{Holmgard2014}, or the need to train an AI via techniques that require very large data sets \cite{Yannakakis2004}. %Though our own research shares these motivations, our primary concern was of the cost of user studies and a need to ensure that the AI driving the adaptations in the user study would be as well tuned as possible before facilitating the study.

\section{Player Modeling via MABs}

% State the problem and why we need to fine-tune the AI prior to the user study
The key idea behind our multi-armed bandit approach is that the MAB strategy serves both as (1) the method by which we model players and (2) the AI that adapts the game to guide player experience in real time. Our approach consists of 3 main components:
\begin{itemize}
	\item The {\bf arms}: the set of possible ways in which the game can be adapted and from which the MAB strategy chooses each time it needs to adapt the game to the player.
	\item The {\bf reward}: the numerical quantity the MAB strategy will aim to maximize, such as the player's daily steps in an exergame designed to encourage walking. %In an interactive fiction game, this could be one of the many functions that have been proposed in the literature of experience management~\cite{weyhrauch1997guiding}.
	\item The {\bf MAB strategy}: the algorithm that chooses an arm, observes the reward resulting from that choice, and updates its internal model of the player to make the next decision.
\end{itemize}

The execution cycle (Figure \ref{fig:motivation-and-steps}) works as follows:
\begin{enumerate}
	\item Initially, the MAB strategy has no information about the player at hand (however, pre-existing individual or population information could be used to initialize the strategy). 
	\item The MAB strategy selects one of the possible arms, and the game is adapted as designated by the arm. %The frequency with which the MAB strategy is asked to select an arm is domain dependent.
	\item The player continues to interact with the game within this adaptation, which results in some measurable metric or metrics that render a ``reward'' value.
	\item The MAB strategy observes this reward and updates its internal model of the player. %The nature of the player model depends on the MAB strategy being used and can range from simple reward averages to much more complex models.
	\item The cycle repeats, and the MAB strategy chooses again.
\end{enumerate}

MAB strategies aim to balance exploration and exploitation, deciding when to {\em exploit} the arm that is currently believed to be the best for the player (according to the objective encoded in the reward function) and when to {\em explore} a different arm in order to learn more about the current player. Therefore, in addition to making the necessary decisions to adapt the game, MAB strategies naturally facilitate methods for obtaining the necessary data to update the player model.

%The following discusses how we instantiate this framework in the context of our application domain.

% 1) How are we creating a player model from an MAB strategy? How are we able to model comparison tendencies as an MAB problem and leverage MAB techniques toward capturing player SCO?
\subsection{Modeling SCO via Multi-armed Bandits}\label{section:modelingSCO}

\begin{figure}[t!]
	\includegraphics[width=\columnwidth]{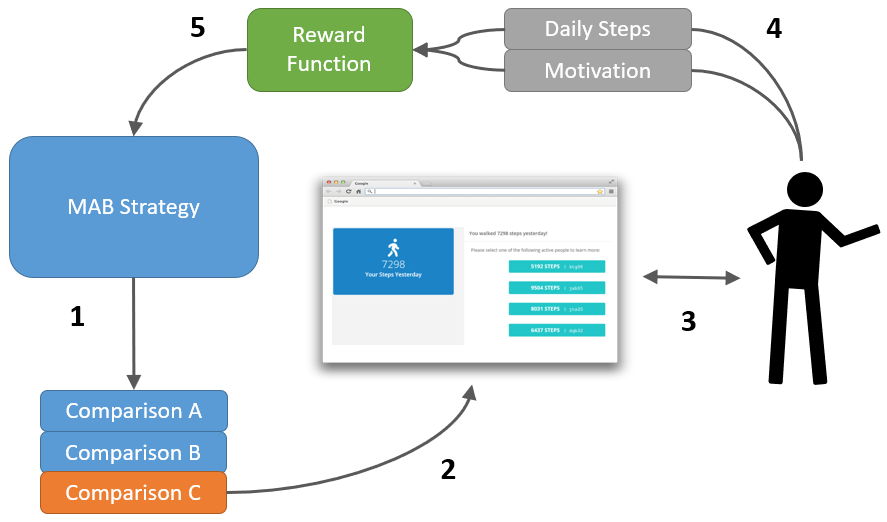}
	\centering
	\caption{Execution cycle for our MAB strategy: 1) MAB strategy selects among multiple configurations of comparisons to present to the player. 2) Comparison options are presented. 3) Player interacts with the software. 4) Metrics are generated (daily steps and self-reported motivation). 5) Reward is observed and recorded by the MAB strategy.}
	\label{fig:motivation-and-steps}
\end{figure}

% Introduce SC/SCO and justify its interest for games researchers
Social comparison is a psychological process in which individuals use comparisons to others, often subconsciously, to assess their degree of success (self-evaluation), to plan for future success (self-improvement), or to view themselves in a favorable light (self-enhancement)~\cite{Wood1989}. Even when objective standards are available, comparisons to salient others can still be preferred and potentially even more influential~\cite{klein1997objective}. 
This carries into gaming, where leveraging social comparison in team competitions has been shown to be effective in influencing participant motivation toward increasing physical activity (PA)~\cite{Zhang2016}.

The details that govern the ways in which a person conducts these comparisons are regarded as individual traits that can be described in aggregate as the person's social comparison orientation (SCO)~\cite{Gibbons1999}, which includes their tendency to perform comparisons, their preference in seeking out targets, and the influence that such comparisons have on their future behavior~\cite{Buunk2007}. Specifically, our research is interested in modeling the degree to which an individual tends to seek out comparison targets performing better than they are (i.e., {\em upward} comparisons) or worse than they are (i.e., {\em downward} comparisons). As discussed later, our simulated players model these features.

% explain our aim regarding MABs, SCO, and player modeling
In our broader research of leveraging motivation psychology toward improving engagement and efficacy of game-related interventions, particularly social 
% TODO: blinded for submission, reinstate for camera-ready
exergames~\cite{caro2018understanding}, 
we seek to model player SCO within adaptive games in a way that can provide dynamic and individualized experiences. %In line with theory regarding SCO~\cite{Gibbons1999} we propose that this preference can be detected as an individual trait through the course of training an MAB strategy that attempts to provide comparison opportunities to that individual with the goal of increasing their motivation toward exercise. 
%Specifically, in a multiplayer game where such comparison opportunities are facilitated and marshalled by the game, we propose that such an MAB strategy could be deployed to implement this player modeling. 
%This strategy, which would be responsible for providing the player with comparison opportunities, would observe player reactions in response to those offerings to eventually predict that preference in each participant.
%
%We theorize that if a player has an innate preference for upward comparisons, then an MAB strategy generating comparison opportunities and observing the player's behavior as a consequence will eventually converge on a strategy that more frequently offers the player upward comparison opportunities. Similarly, a preference for downward comparison will be exposed by an MAB's convergence toward providing downward comparisons.
%This MAB can then be used in an EM system designed to alter or optimize the gaming experience.  In this way, we propose that the MAB itself can implement a player model capable of facilitating an adaptive game that promotes enjoyment and motivation to play.
%
%What remains is a method for codifying player comparison opportunities as electable arms in an MAB strategy. 
This paper is a first step in this direction, where we evaluated our SCO player modeling approach in a simpler web-based intervention that gave players an opportunity to log in and compare themselves against other profiles. In our application, we instantiated the three elements of the MAB player modeling framework as follows (Figure \ref{fig:motivation-and-steps}):
\begin{itemize}
	\item {\bf Arms:} The MAB strategy had an opportunity each day to choose which comparison opportunities were displayed. Our setup had 3 arms: arm ``A'' presented the player with zero upward comparison opportunities (i.e., all other displayed profiles walked fewer steps than the player); arm ``B'' presented the player with two upward and two downward, and arm ``C'' offered the player four upward comparisons. It is expected that a player's act of comparing themselves to these profiles, depending on their individual SCO, would result in a change in motivation. Once the configuration was chosen, profiles were presented, and the player was given an opportunity to investigate more details of only one of the profiles.
	\item {\bf Reward:} The player's eventual steps $s$ that day following the session as well as a self-reported motivation score $m$ on a 5-point Likert scale following the session (players reported their motivation before the session as well) were used to calculate a reward score $r_t$ using the following formula (where $\mu$ and $\sigma$ represent mean and standard deviation with respect to all previously observed data for that player): %On the following day, the MAB strategy incorporates both the step and motivation data to drive future decisions. As this process is repeated each day, the MAB strategy eventually converges on the type of comparison that maximizes a reward score $r_t$ calculated from both steps $s$ and reported motivation $m$ according to the following formula, averaging the values of both after standardizing them relative to all observations so far for that player:
	\[ r_t = \frac{\frac{s_t - \mu_s}{\sigma_s}+\frac{m_t - \mu_m}{\sigma_m}}{2}\]
	
	\item {\bf MAB strategy:} we evaluated a large collection of strategies enumerated in Section~\ref{mab-strategy-assessment}. 
\end{itemize}

The next section describes the approach we used to evaluate the different MAB strategies and parameters by using simulated players. As detailed later in the paper, we then evaluated the best performing strategy with real players.

%The remainder of this section describes our approach to design and evaluate our MAB approach before conducting user studies with real users.

% 2) How do we create simulated players that can maintain enough parity with real players with respect to SC behavior?
\section{Simulated Players}

The purpose of creating simulated players was to evaluate different MAB strategies while modeling players that exhibit similar statistical trends as real users (i.e., same variance in numbers of steps per day). This was crucial in our case, as it was unclear whether any MAB strategy would converge fast enough given the expected duration of the study and the large degree of noise present in real human data.
Our simulated players had three main components:
\begin{enumerate}
	\item Step Model: A probabilistic model that simulated the number of steps typical humans take in a day.
	\item SCO Data Model: A representation of a player's tendency toward upward and downward comparisons.
	\item SCO Behavioral Model: A set of functions implementing player behavior given the step model, the SCO data model, and the player's social comparison activities.
\end{enumerate}

\subsection{Step Model}\label{step-model}
\begin{figure}[t!]
	\includegraphics[width=\columnwidth]{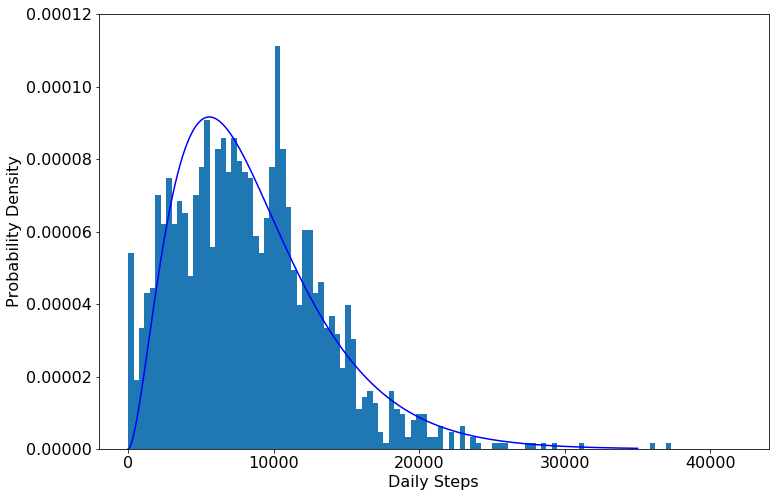}
	\centering
	\caption{Daily step data from the Mechanical Turk experiment \cite{Furberg2016}. Histogram is overlayed with a probability density function curve for a Gamma distribution with k=2.8, $\theta$=3100. } \label{fig:mechanical-turk-gamma}
\end{figure}

In order to obtain a realistic step model, and in consideration for the bias-variance tradeoff discussed by Carterette et al. \cite{Carterette2011}, we opted to leverage existing behavioral data. Specifically, we obtained data from a publicly available Mechanical Turk survey conducted over three months in 2016 by Furberg et al. \cite{Furberg2016}. After omitting days with zero steps, we confirmed via D'Agostino-Pearson and Shapiro-Wilk tests that the data was not from a normal distribution (both $p < 0.01$).
%
%Though we would expect high variation based on location and lifestyle, 
Previous research has suggested human walking patterns align with gamma distributions \cite{Orendurff2008}, reflecting a common trend among intermittent human behaviors \cite{Guo2011}, which we used to fit the data (Figure \ref{fig:mechanical-turk-gamma}).  %Notice that our goal was not to confirm that human steps follow a specific distribution, but rather to find a model that we can use to produce random values in our simulation that over time will produce a distribution that sufficiently approximates that of the values we observed in real human data. With our estimated distribution defined, 
%This gave our simulated players the ability to walk each day according to this distribution. This is important, as we wanted to evaluate how the MAB strategies would learn when faced with data distributed in this particular way.

\

\subsection{SCO Data Model}\label{section:sco-data-model}
%We then attempted to codify aspects of individual SCO traits that reflect the current psychology models regarding apsects of SCO. Namely, 
We considered the following SCO traits for the simulated players:
%, where these components have been proposed as aspects of SCO that explain the way an individual will react to a comparison event:

\begin{enumerate}
	\item Direction: A propensity for a player to more often make (deliberately or subconsciously) upward or downward comparisons. This is referred to as the player's directional {\em preference} for social comparison~\cite{Wood1989}.
	\item Intensity: The general degree of influence that SCO activities have in a simulated player's motivation and behavior.
	%\item Valence: {\color{blue} REWRITE: A tendency toward positive or negative affect changes when faced with either comparison activity}
	%{\color{red} [santi: I do not understand this point neither here nor below (marked both places in blue). What do you mean?]}
\end{enumerate}

To achieve this, two parameters $0 \leq u \leq 1$, $0 \leq d \leq 1$ were used that represent the simulated player's affinity toward upward and downward comparisons on a linear scale. This was chosen as two separate variables to reflect the design of the common psychology instrument used to measure SCO--namely, the upward and downward comparison subscales of the Iowa-Netherlands Comparison Orientation Measure (INCOM) \cite{Gibbons1999}. 

%In our simulated players's SCO model, 
The propensity to prefer one comparison over another (1) is modeled as the proportion defined by the simulated player's $u$ and $d$ values. E.g., an assignment of $(0.8,0.4)$ would indicate a 2x preference toward upward comparisons.
A simulated player's general sensitivity to either comparison (2) is modeled by the magnitude of the value. E.g., an assignment of $(0.0,0.5)$ would designate a simulated player not influenced at all by upward comparisons but moderately influenced by downward comparisons.

%{\color{blue} REWRITE: As for a player's tendency toward positive or negative affect changes when faced with comparisons (3), this is not addressed by the INCOM, and a direct corollary did not exist for our $(u,d)$ variables. Therefore, we chose to assume a notion of aversion (negative affect) toward a comparison by negating the opposing value ($u$ or $d$). In other words, just as $u$ represents preference for upward comparisons, we model $-u$ to represents aversion to downward comparisons.}
%{\color{red} [santi: you earlier said that d and u are between 0 and 1, but now we have negatives. So, they are between -1 and 1 then, right? Also, I do not really understand (3), but if it's what I kind of think it means. Probably separating (2) and (3) makes it more confusing, and we can just say that u and d control how much each comparison affects the simulated player behavior]}

\subsection{SCO Behavioral Model}
The simulated players were given a programmatic version of the same exercise intended for real human users in an upcoming user study. The details of this exercise are explained in further detail in Section~\ref{section:methodology}, and involve a repeated interaction over the course of 21 days (i.e., time steps). In each time step, the simulated player is given a list of four {\em profiles} depicting the PA behavior and other details for four realistic (but fabricated) people. The PA performance of these profiles would be strategically generated to provide upward or downward comparisons for the player, according to the simulated player's own steps the previous day and the MAB strategy's assessment of their preference for social comparison. The simulated player then chooses to view one of the profiles in detail, and (presumably influenced by that experience of comparing their PA output to that of another) afterward generates a value for their ``steps'' that day.

The decisions made in this process and the value of the generated steps were influenced by the simulated player's internal SCO data model via the behavioral models described below. Specifically, each simulated player was equipped with three behavior models: {\em selector}, {\em step simulator}, and {\em motivation}.

%The first was a selector that, given a set of player profiles to compare against, would choose one of them based on $(u,d)$. The second is a step simulator that uses both the step model and the SCO data model. The third is a method for the simulated player to self-report motivation before and after the intervention. %We describe these below.

The {\it selector} component considers the list of the four potential player profiles for comparison and chooses one of them.
% To do so, it leverages the first aspect of SCO (1) conveyed in the SCO Data Model regarding propensity to choose upward or downward comparisons.  
The choice (resulting in a {\em comparison target} for that day) is determined by the simulated user's underlying $(u,d)$ values, where a direction preference is stochastically selected, weighted by $u$ and $d$.  E.g., if the simulated player had values $(0.4,0.2)$, they would be twice as likely to choose an upward comparison than a downward comparison. Once the direction is determined, the user selects randomly among the choices available in that direction. If no choices are available in that direction (e.g., the simulated user chose to select downward today but the MAB strategy chose Arm C and provided four upward comparisons), a random choice is made from the remainder.

The {\it step simulator} component reports the simulated user's daily steps following this selection and resulting comparison event. %To do so, it leverages the second aspect of SCO (2) conveyed in the SCO Data Model regarding degree of influence that comparisons have over the simulated user. 
%To do so, the step model is queried to sample a number of steps $s'_t$ from the gamma distribution. This is further modified by the relative magnitude of the simulated player's performance the previous day $s_{t-1}$ (number of steps the previous day) and the comparison target's performance, $s^t_{t-1}$ as well as $(u,d)$. Specifically, the number of steps $s_t$ reported for an upward comparison is:
%
%\[s_t = s'_t \left(1 + u \frac{s^t_{t-1}-s_{t-1}}{s_{t-1}}\right)\]
%
%For a downward comparison it is:
%
%\[s_t = s'_t \left(1 + d \frac{s_{t-1}-s^t_{t-1}}{s_{t-1}}\right)\]
%
%For example, consider the steps reported by a simulated player with $u=0.4$ following a comparison to another player achieving 12\% more than them (i.e., upward comparison). A value is randomly pulled from the gamma distribution to yield $9174$ steps. The final daily steps returned for that user would be $s_t = 9174 * (1 + (0.4 * 0.12)) = 9614$ steps. 
To do so, the step model is queried to sample a number of steps $s'_t$ from the gamma distribution discussed in Section~\ref{step-model}. This is further modified by the comparison target's performance, $s^c_t$ as well as $(u,d)$. Specifically, the number of steps $s_t$ reported for an upward comparison is:
\[s_t = s'_t \left(1 + u \frac{s^c_t-s'_t}{s'_t}\right)\]
and
\[s_t = s'_t \left(1 + d \frac{s'_t-s^c_t}{s'_t}\right)\]
for a downward comparison.

The {\it motivation} component enables the simulated player to self-report their motivation both before and after a comparison.  Because no public data existed for user motivation reporting, % on our Likert scale question (ranged 1 to 5), we presumed most users would self-report a pre-comparison motivation level of 2, 3, or 4.
our simulated players select uniformly at random from values 2, 3, and 4 in the 5-point custom Likert scale used for self-reporting motivation. Motivation reported after the comparison is determined by an aggregate affect value calculated as $u - d$ for an upward comparison and $d - u$ for a downward comparison. %To simulate motivation change, we leverage the third aspect of SCO (3) conveyed in the SCO data model regarding the influence that comparisons of either type respectively have on a simulated player's affect. 
%The $u$ or $d$ value corresponding to the nature of the comparison activity (modeling attraction and positive affect) is added to the negative of the opposite value (modeling aversion and negative effect) to achieve an aggregate affect value
%
If this aggregate affect value is positive, then a motivation score higher than or equal to the initial value is randomly selected; if it is negative, than a motivation score lower than or equal to the initial value is randomly selected; and if it is zero, then a motivation score equal to or adjacent to (higher and lower) the initial motivation is randomly selected.
%For example, if a simulated player assigned $(0.4,0.1)$ encountered a downward comparison, they would be assigned an aggregate affect of $0.1-0.4=-0.3$. Because this value is negative, if their initial motivation was $3$, their reported motivation following that comparison would be randomly selected from the set of $[1, 2, 3]$.

% 3) how can we use simulation to fine-tune our SCO MAB before the user study? What experiments did we run? What did we ultimately decide on?
\subsection{MAB Strategies}\label{mab-strategy-assessment}

We conducted experiments on our simulation reflecting the user flow of the anticipated user study, using simulated players (instead of real players) and multiple MAB strategy variants. %After requesting initial motivation from the simulated players, the MAB strategy chooses an arm that determines which other four player profiles will be shown to the player for potential comparison. The simulated player chooses to compare to one of the  profiles from this selection and then reports daily steps and motivation following that comparison. The MAB receives this information, completing one interaction episode. %This cycle repeats, where the MAB strategy uses results from previous episodes to adjust its configuration selection for the player in subsequent episodes.
Specifically, we compared the following MAB strategies:
%The three arms the MAB strategy chooses from have the following effects: arm ``A'' offers four downward comparisons, arm ``B'' offers two upward and two downward, and arm ``C'' offers four upward comparisons. We compared the following MAB strategies: 

\begin{itemize}
	\item Random: used as a baseline strategy for evaluation where the arms are always selected at random.
	\item UCB1~\cite{auer2002finite}: calculates a score for each arm based on past average reward and a confidence interval (arms selected less often have less data and therefore lower confidence in expected value). UCB1 balances exploiting arms with proven rewards and exploring arms not yet selected enough to create tight confidence intervals.
	\item $\epsilon$-greedy: selects the best historically performing arm except for a certain percentage of the time (designated by $\epsilon$) where the strategy will randomly explore another arm.
	\item $\epsilon$-first: selects randomly at the beginning of the experiment until a point specified by $\epsilon$, after which the best historically performing arm is always selected.
	\item $\epsilon$-decreasing with linear decay: similar to $\epsilon$-greedy, except that the $\epsilon$ parameter starts higher and gradually decreases to a lower value over a specified number of steps.
	\item $\epsilon$-decreasing with exponential decay: an $\epsilon$-decreasing implementation that decreases the $\epsilon$ factor according to a specified exponential decay curve rather than a linear schedule.
\end{itemize}

\subsubsection{Regression Strategy Variants}\label{section:regression-variants}

Each of the MAB strategies listed above maintains an internal expectation of the reward for each arm, calculated as the average of all rewards previously reported when selecting that arm. Since the amount of variance we observed in the publicly available data was very large, we designed another set of strategies equipped with better score estimation techniques.

Specifically, we used linear regression involving reported steps and motivation over the previous days to predict the expected score that would be obtained when selecting a given arm in the next decision. We are aware that such a modification challenges the status of these approaches as MAB strategies, because MAB strategies are stateless; rather, the introduction of reward prediciton via linear regression aligns more with a general reinforcement learning paradigm. But for simplicity, we present these strategy variants as modified MAB strategies.

%The above strategies were initially tested on either steps, motivation responses, or some arithmetic combination of both. However, this {\it reward calculation} would have a large impact on the operation of the MAB strategies, and a critical task was to determine the appropriate form of the steps and motivation feedback to use to drive the MAB's selections for this scenario. We therefore decided to explore beyond traditional reward averages and pursue MAB predictions based on multiple regression models for both steps and self-reported motivation levels.

%To maximize the effectiveness of our step-based regression model, we again leveraged the Mechanical Turk experiment data to inform our feature set. Our starting model for predicting baseline daily step behavior on a given day for a given person included that person's steps for each of the previous 7 days as well as the day of the week at that time.

Running linear regressions for each day in the Mechanical Turk data set and observing $p$-values for the correlations of each feature, we performed iterative backward elimination to ultimately end with a model in which all features correlated with a significance of $p < 0.05$.  Our final model for predicting a person's daily steps from historical data consisted of their steps 1, 2, 3, 4, 6, and 7 days prior, as well as whether the day of the week was Monday or Friday. Because no such data existed for motivation, we chose an initial approach of building our regression model from the three previously reported motivation values.

We incorporated this regression model into the $\epsilon$-based strategies, where we replaced the {\em mean} operation with this linear regression, resulting in ``regression-based'' variants for each of them (e.g., ``regression-based $\epsilon$-decreasing with exponential decay'').

\section{Results}

In this section, we examine both the results of our simulator runs to prepare for the user study and the results of the user study deploying the MAB strategy that worked best in simulation.

\begin{figure}[t!]
	\includegraphics[width=\columnwidth]{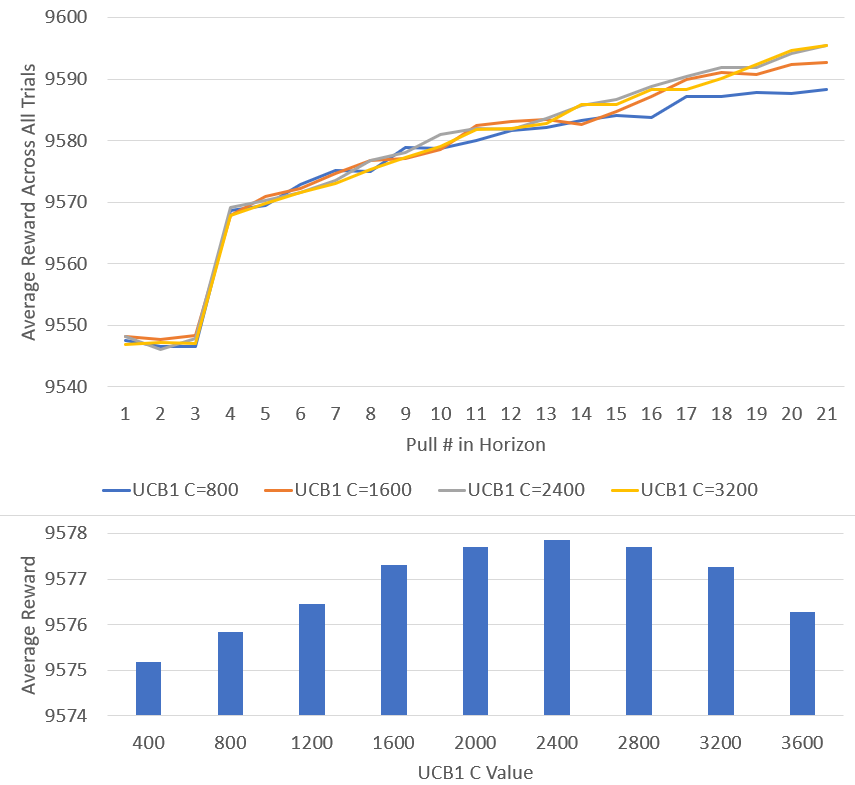}
	\centering
	\caption{Average reward (vertical axis) obtained using a UCB1 MAB strategy for different values of $C$ over time (horizontal axis, top), and global averages (bottom).}
	%Tests of best $C$ Value for the UCB1 strategy. UCB1 is most effective when a particular parameter $C$ is tuned to a magnitude most appropriate for the rewards of a simulation. This experiment tested several UCB1 strategies with different values for $C$ (400 to 3600) at 50 million trials each. The top chart shows average performance over the horizon. The bottom chart adds clarity by graphing the average reward for each variant across all pulls and trials, suggesting that $C=2400$ is the closest to the optimal value for $C$.}
	\label{fig:experiment1}
\end{figure}

\begin{figure}[t!]
	\includegraphics[width=\columnwidth]{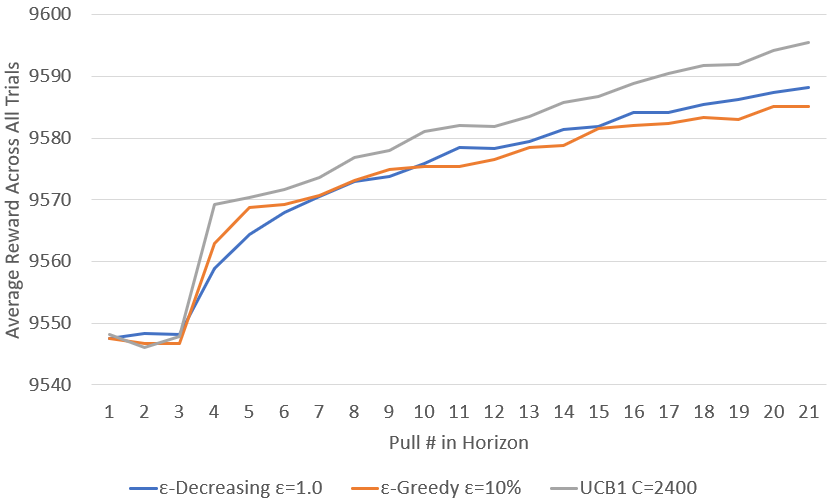}
	\centering
	\caption{Average reward (vertical axis) for three different MAB strategies (UCB1, $\epsilon$-greedy, and $\epsilon$-decreasing) over time (horizontal axis).}
	%Examining several promising MAB strategies via the simulator. Results from this run (50 million trials) suggested continuing investigations into both UCB1 and $\epsilon$-decreasing in preparation for our user study.}
	\label{fig:experiment2}
\end{figure}
	
\subsection{Results of Simulations}

%Explain the results from our SQ2.1-based player simulator and how we worked to discover an effective strategy before the study. Present some examples of experiments between strategies and parameters and how these findings helped shape toward our final approach.
%The final task prior to the user study was to run experiments to determine the best MAB strategy and configuration from the large collection of possibilities available that would result in the highest motivational results for our simulated players, thereby informing the strategy to pursue in our real user study.

We performed three sets of experiments, where in each a specific MAB configuration was tested with a given simulated player. Each experiment conducted $N$ experimental trials, where an experimental trial consisted of the simulated player interacting with the MAB strategy over $M$ steps (simulated days). In each step, the MAB strategy was queried for its decision, which would be given to the simulated player, and the player would simulate behavior in response (e.g., selecting one of the four presented profiles and generating resulting steps for that day). This would be reported to the MAB strategy, which would update its internal statistics. The MAB state was reset at the beginning of each trial. 
We report the average number of steps performed by the simulated players each of the simulated days over all the experimental trials. We used $u = 0.3$ and $d=0.6$ for all the simulated players in our experiments (estimating the ranges of $u$ and $d$ values that most resemble human behavior is part of our future work).

%statistics for results of each simulated day across all trials would be computed, such as averages, standard deviations, confidence intervals, and histograms of arm selections.  These would be exported by the simulator, where they could then be used to create graphs like those in Figures \ref{fig:experiment1}, \ref{fig:experiment2}, and \ref{fig:experiment3}. The following are three such experiments conducted to explore the performance of these strategies.

% {\color{blue} one figure per example, and we will explain exactly how many iterations were run in each experiment, which were exactly the MAB strategies/parameters compared in each experiment, and then a small explanation of the results, listing which one worked the best.}

% Experiment 1
\subsubsection{UCB1 C-Value Experiment}

%Before we could examine performance across MAB strategies, 
Some strategies needed to fine-tune their parameters to the specific task, such as the $\epsilon$ parameter in $\epsilon$-greedy strategies or the $C$ parameter in UCB1. 
%For example, in the case of the $\epsilon$-greedy strategy, successful parameters around the the threshold for exploration (defined by $\epsilon$) need to be determined before that strategy can be compared to others like UCB1. Similarly, UCB1 relies on a parameter $C$ that allows it to normalize rewards by anticipating the range of values it should be expected to receive. 
In our simulation experiment depicted in Figure \ref{fig:experiment1}, we report on a set of runs that set out to to tune the $C$ parameter for the UCB1 strategy. Notice the uncommonly high values for $C$, as a result of the fact that the reward function also returns very high values (far from the usual $[0-1]$ interval).

In this experiment, we ran $N=50$ million trials of UCB1 for $C=k*400$ for $k \in \{1,...,9\}$. The experiments targeted a horizon of $M=21$ steps in anticipation of a user study lasting 21 days. Our UCB1 implementation requires that every arm be evaluated one time before the strategy engages, which is the cause for the lower results in the first three pulls. From this data, it appears that $C=2400$ achieved the largest reward in our experimental setup.

In the interest of space, we do not report the detailed results for tuning $\epsilon$-greedy strategies, but the best parameters were found to be $\epsilon = 0.1$ for $\epsilon$-greedy, and $\epsilon = 1.0$ for $\epsilon$ decreasing.

% Experiment 2
\subsubsection{Strategy Comparison Experiment}

In this experiment, we compared three of the most promising strategies to determine the top candidates to investigate for deployment in our user study.

Figure \ref{fig:experiment2} shows the results of $N=50$ million trials for each of UCB1 ($C=2400$), $\epsilon$-decreasing  ($\epsilon=1.0$), and $\epsilon$-greedy ($\epsilon=0.1$). As before, we targeted a horizon of $M=21$ steps with a requirement that all test each arm once (steps 1-3) before engaging their strategy (forced exploration). The results suggested an advantage in the UCB1 strategy over the others. However, of the two $\epsilon$-class strategies, $\epsilon$-decreasing appeared to perform the best.

% Experiment 3
\subsubsection{Regression-Based Experiment}

\begin{figure}[t!]
	\includegraphics[width=\columnwidth]{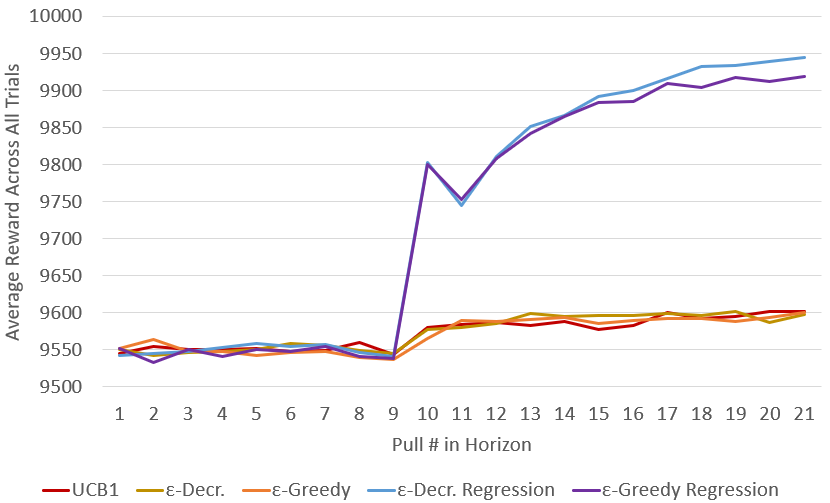}
	\centering
	\caption{Average reward (vertical axis) for three different MAB strategies over time (horizontal axis) with a nine-step forced exploration period, comparing strategies with linear regression and without.}
	%Following the forced exploration period, regression variants of the strategies (blue shaded lines) generally outperformed their reward average-based counterparts (red and yellow shaded lines). }
	\label{fig:experiment3}
\end{figure}

%As our design for the user study finalized, we were able to also finalize the design of our regression models to serve as an evaluation function within the MAB strategies as an alternative to a simple average of rewards. We were able to also use our simulator to test this approach, comparing regression variants of our existing strategies against each other. 
The last experiment compared our best performing strategies from Experiment 2 to regression-based strategies. Figure \ref{fig:experiment3} graphs the results of the experiment with $N=1$ million trials. In this experiment, we also introduced a nine-step {\em forced exploration} period in which each of the three arms were pulled three times (in random order) before the MAB strategy was engaged, an approach that has been found to be advantageous in short-horizon MAB scenarios~\cite{gray2020regression}. Our results showed that the regression variants performed significantly better than the non-regression models (experiments comparing different forced exploration periods are not reported in the interest of space).

Results from simulated experiments ultimately led to our selection of MAB strategy for the user study: an $\epsilon$-decreasing strategy implemented with an exponential decay of $1/x^\epsilon$, an $\epsilon$ value of 1.0, and a nine-step forced exploration period.

%The breadth of our experimentation explored many classic MAB strategies, variants of those strategies (such as $\epsilon$-decreasing with linear decay versus exponential decay), parameter options for those strategies, the effect of forced exploration (i.e., compelling the MAB to explore for a certain period before engaging strategy), Combinatorial (CMAB) variants of our traditional strategies, CMAB hybrids that implemented na\"{i}ve strategies \cite{Ontanon2017}, the use of multiple regression analysis versus typical average-reward valuation, and other aspects relevant to our AI's construction. 

\subsection{Results with Real Users}\label{section:studyvalidation}

%\begin{figure}[t!]
%\includegraphics[width=\columnwidth]{img/SQ21website.png}
%\centering
%\caption{Web-based fitness activity presents players with opportunities to select profiles that facilitate upward and downward comparisons.} 
%\label{fig:sq21-website}
%\end{figure}

Finally, we evaluated the best performing MAB strategy from the simulated experiments with real users via a 3-week study. Although the long-term goal of this project is to design a full-fledged game with this technology, we limited ourselves in this study to a web-based activity as a first step toward that goal.

%We then sought to deploy our tuned MAB as the Experience Manager in an interventional user study regarding perceptions of lifestyle physical activity (PA). In particular, this study aimed to view the potential effect that social comparison events have on a person's daily walking activity (measured in steps) and self-reported motivation to exercise (measured via self-reported survey). Over the course of the 3-week study, the MAB's role was to classify participants on aspects of SCO and aim to provide social comparison opportunities that aligned with their determined preferences. The MAB-focused results of the this study aimed to show a higher change in daily steps and higher self-reported motivation in an experimental group with social comparison opportunities curated by the MAB versus a control group in which users engaged the exact same activities while being offered opportunities at random.

%It is intended that similar social comparison activities could then be integrated into gameplay or other software workflows (such as in an interactive leaderboard), where a similarly tuned MAB could be leveraged for game adaptation or other AI-driven action.

\subsubsection{Methodology}\label{section:methodology}

The participants were recruited from psychology and digital media courses at Drexel University, where they were informed they would be participating in a study regarding attitudes toward health. They were set up with pedometers (i.e., smartphones equipped with accelerometers and Fitbit software) to track their daily steps and were then directed to engage in daily sessions with a web-based software application. It was requested from each participant that they complete one session (around 5 min.) each day for 21 days, which consisted of the following:

After logging in, participants were asked to rank their motivation to exercise on a scale from ``very low'' (1) to ``very high'' (5), after which they were presented with their own step count from the previous day.  They were then shown four buttons representing profiles of other people that they could investigate. These profiles were created by the research team and did not represent real people, but they were presented as real and the participants were not informed that they were fabricated. % The buttons offered minimal information regarding each profile, specifically an abstract username (e.g., ``jts20'') and step count for the previous day. 
Participants were requested to select a profile among the four options to view additional details regarding that profile beyond simple step count (e.g., diet, hobbies, exercise habits, profession, etc.).
%
%Because the only comparable information offered in the initial choice was the step counts presented for each of the profiles, it was intended that the step count would be the driving factor in the participant's daily decision.  %As the social comparison component of this study, the step counts offered in these profiles were specifically generated each day to span a range around the steps of the participant. 
The MAB strategy's choice %for configuration (i.e., downward-only, 2-of-each, and upward-only) 
would dictate which profiles would be given in order to offer those comparisons. %For example, the downward-only would offer four profiles that were 60-90\% that of the participant's own steps, while the upward would offer four profiles with values 110-140\% of the participant's step count.
After the players were done inspecting the selected profile, they were asked again to report their motivation to exercise.
Participants were divided into two conditions: experimental (with MAB strategy engaged) and a control condition (that received random arm selections). %Notice that since the best performing MAB strategy had a forced exploration period of 9 iterations, during the first 9 days, participants on both groups received basically a similar experience, and it wasn't until day 10 that the MAB started to make a difference. %After a 9-day period of forced exploration where each of the participants was given an equal amount of all arms, the AI implemented the respective strategy for each participant for the remainder of the study. Following each session, the self-reported motivation data and daily steps for each participant was recorded by the MAB strategy (regression-based $\epsilon$-decreasing), which it used to inform future decisions.

\subsubsection{Results}

% Number of people and discuss the evaluation we're interested in
A total of 53 people enrolled in the study, but five participants did not finish enough sessions to qualify as having completed the study (at least 14 days). Of the remaining 48 participants, 25 were in the control condition and 23 were in the experimental condition. %For the purpose of MAB evaluation, we are interested in an observable steps and motivation differences between the two conditions on days in which the participants were potentially influenced by their social comparison activity prescribed by the MAB (or otherwise prescribed randomly) during their session.

% Note about Fitbit sync failures on 11/01-11/03
% [santi: I commented this out, too much detail for this paper, we will discuss this in the full paper of this study]
%As a limitation for the study, there was a three-day period in which our web application was not able to fetch accurate values for participant steps from the Fitbit servers. Though participant sessions continued during this time, this issue did not affect participants equally. Therefore, data from these three days have been omitted from the analysis.

\begin{table}[t!]
	\centering
	\caption{Difference in step counts (between previous day and current day) and reported motivation (before and after session) during the intervention period for participants with and without the MAB strategy (2-tailed T-test, $\alpha$=0.05).}
	\label{tbl:userstudy}
	\begin{tabular}{|c|c|c|} \hline
		Condition & Steps change & Motivation change \\ \hline
		Control  & 42 & 0.013 \\
		Experimental     & 160 & 0.111 \\ \hline
		T-score: & 0.3007 & 1.9908 \\ 
		p-value: & 0.764 & 0.047 \\ \hline
	\end{tabular}
\end{table}

% Evaluation 1 - did the exp. condition drive a higher change in steps?
Results are shown on Table \ref{tbl:userstudy}, where we found that participants in the control group saw an average of 42 extra steps on the days of their sessions (with respect to the day before) compared to 160 extra steps for participants in the experimental group. Though perhaps representing a trend, this finding was not found to be statistically significant ($p$=0.764) via two-sample T-test at $\alpha=0.05$ (two-tailed, dof=445).
% Evaluation 2 - did the exp. condition drive a higher change in motivation?
%Next, we examined the average change in self-reported motivation to exercise (difference between pre-session value and post-session value) as an immediate result of completing the social comparison activity of the session.
However, the change in motivation before and after inspecting the selected user profile did demonstrate a statistically significant difference ($p$=0.047) via two-sample T-test at $\alpha=0.05$ (two-tailed, dof=388), where participants in the control group saw an average motivation score increase of 0.013 compared to an increase of 0.111 in the experimental group.

\section{Discussion}

Though metrics such as daily steps and player motivation might be affected by many factors, the increase in steps and the statistically significant increase in self-reported motivation suggest that our bandit-driven manipulation of selecting individualized social comparison targets for users was more effective than random assignment. This in turn appears to support our approach of defining our player modeling problem as an MAB problem, to which we were able to apply a wealth of theory and solutions already developed by that field. To our knowledge, this is the first case in which an MAB-based approach has been applied toward player modeling in order to implement an adaptive game.
%Further, we believe it to be the first instance of an adaptive interactive experience that bases its model on the player's social comparison orientation (SCO). RCG: See Klein-2017, Mollee-2016

In our case, this technique for player modeling via MAB strategies allowed us to engage in both a top-down and bottom-up approach to player modeling simultaneously~\cite{Yannakakis2013}. Theory borrowed from the psychology field of social comparison helped to define the arms for our MAB problem (i.e., enjoying the theory-based insight inherent in top-down modeling), while avoiding the need to provision our classification system with context-specific heuristics to define players (i.e., the advantages of bottom-up modeling). Rather, in our case the mechanism is the model, and the context-agnostic operation of the MAB strategy allowed the system to assess players based simply on a reinforcement loop (i.e., user response in terms of steps and self-reported motivation) instead of the researcher's perceptions or interpretations of player behavior.

Further, these results support the proposed benefits of the technique of implementing an AI-based intervention first as a simulation in order to explore the potential options for the AI. In this practice, simulated users were constructed with data and behavioral models (based on psychology theory) that allowed them to exhibit behaviors on which we conducted multiple experiments. The results of these experiments, which were achieved with greater speed and lower cost than preliminary user studies, informed our decisions prior to recruiting human players for our planned user study. 

\section{Conclusion}

% SUMMARY: in 3-4 lines summarize what you presented, e.g. "This paper presented a new approach to blah blah"
This paper presented a new approach to player modeling based on multi-armed bandits (MABs). MABs naturally model both the problem of exposing the player to different situations to build an accurate player model and the problem of adapting a game to maximize features of interest to the designer. We also presented a method for creating simulated players to evaluate and fine-tune these MAB techniques before deploying with real users. %This is an important step, as the number of interactions with the player we can expect in some user studies might be much lower than the numbers typically used in the literature to evaluate MAB strategies.

Our results indicated that an $\epsilon$-decreasing strategy with a nine-step forced exploration period and a linear regression model to estimate both steps and motivation performed the best in simulation and therefore was used with real users. Our results showed the difference in step count increment with respect to the previous day and motivation change were both higher for the experimental condition using the MAB, although only the latter was found to be statistically significant. This is no easy task, as the degree of variance in both step and motivation data is very high, and the MAB was able to select arms that achieved positive results in just 21 interactions with the users.

As part of our future work, we plan to improve our simulated user framework to obtain more realistic user behavior models. We also plan to investigate more sophisticated MAB approaches such as contextual bandits~\cite{chu2011contextual} or combinatorial bandits~\cite{Ontanon2017}, which would allow us to integrate state knowledge or engage complex decisions. We are also interested in comparing the estimations built by the MAB strategy with results obtained from standard psychological SCO tests to measure agreement. Finally, our next step is to incorporate our new approach into our game prototype.

\section{Acknowledgements} 
This work is partially supported by the National Science Foundation under Grant Number IIS-1816470. The authors would like to thank the participants of our user study and all current and past members of this project. Special thanks to Jennifer Villareale and Diane Dallal for facilitating the user study and user data collection which is used in this paper.

% The next two lines define the bibliography style to be used, and the bibliography file.
\bibliographystyle{ACM-Reference-Format}


\end{document}